\documentclass[11pt,a4paper]{article}
\usepackage[hyperref]{eacl2021}
\usepackage{times}
\usepackage{latexsym}
\usepackage{lipsum}
\usepackage{longtable}
\usepackage{booktabs}
\usepackage{graphicx}
\usepackage{amsmath}
\usepackage{subfiles}

\usepackage{microtype}

\aclfinalcopy % Uncomment this line for the final submission
\title{The Effectiveness of Morphology-aware Segmentation in Low-Resource Neural Machine Translation}

\author{Jonne Sälevä \\
  Brandeis University \\
  \texttt{jonnesaleva@brandeis.edu} \\\And
  Constantine Lignos \\
  Brandeis University \\
  \texttt{lignos@brandeis.edu} \\}

\date{}

\begin{document}
\maketitle
\begin{abstract}
This paper evaluates the performance of several modern subword segmentation methods in a low-resource neural machine translation setting.
We compare segmentations produced by applying BPE at the token or sentence level with morphologically-based segmentations from LMVR and MORSEL.
We evaluate translation tasks between English and each of Nepali, Sinhala, and Kazakh, and predict that using morphologically-based segmentation methods would lead to better performance in this setting.
However, comparing to BPE, we find that no consistent and reliable differences emerge between the segmentation methods. 
While morphologically-based methods outperform BPE in a few cases, what performs best tends to vary across tasks, and the performance of segmentation methods is often statistically indistinguishable.
\end{abstract}

\section{Introduction}

Despite the advances of neural machine translation (NMT), building effective translation systems for lower-resourced and morphologically rich languages remains a challenging process. 
The lack of large training data sets tends to lead to problems of vocabulary sparsity, a problem exacerbated by the combinatorial explosion of permissible surface forms commonly encountered when working with morphologically rich languages.

Current NMT systems typically operate at the level of \emph{subwords}. 
Most commonly, these systems achieve vocabulary reduction by decomposing tokens into character sequences constructed by maximizing an information-theoretic compression criterion. 
The most widely used subword segmentation method is byte pair encoding, originally invented in the data compression literature by \citet{gage1994new}, and introduced to the MT community by \citet{sennrich-etal-2016-neural}.
Another approach to open vocabulary NMT has been to compose characters or character n-grams to form word representations \citep{ataman2018compositional,ling2015character}.

As BPE has become mainstream, the question of whether segmenting words in a linguistically-informed fashion provides a benefit remains open. Intuitively, the translation task may be easier when using subwords that contain maximal linguistic signal, as opposed to heuristically derived units based on data compression.
The greatest benefit may come in low-resource settings, where the training data is small and biases toward morphological structure may lead to more reusable units.

We seek to address this question by exploring the usefulness of linguistically-motivated subword segmentation methods in NMT, as measured against a BPE baseline.
Specifically, we investigate the effectiveness of morphology-based segmentation algorithms of \citet{ataman2017linguistically} and \citet{lignos2010learning} as alternatives to BPE at the word or sentence level and find that they do not lead to reliable improvements under our experimental conditions. We perform our evaluation using both BLEU \citep{papineni-etal-2002-bleu} and CHRF3 \citep{popovic-2015-chrf}.
In our low-resource NMT setting, all these methods provide comparable results.

The contribution of this work is that it provides insights into the performance of these segmentation methods using a thorough experimental paradigm in a highly replicable environment.
We evaluate without the many possible confounds related to back-translation and other processes used in state-of-the-art NMT systems, focusing on the performance of a straightforward Transformer-based system. To analyze the performance differences between the various segmentation strategies, we utilize a Bayesian linear model as well as nonparametric hypothesis tests.

\begin{table*}[tb]
\centering
\begin{tabular}{llrrr}
\toprule
Translation task &  Split &  Sentences & Tokens (EN) & Tokens (non-EN) \\
\midrule
        NE $\leftrightarrow$ EN &  Train &     563,947 & 4,483,440 & 4,200,818 \\
        SI $\leftrightarrow$ EN &  Train &     646,781 & 4,837,496 & 4,180,520 \\
        KK $\leftrightarrow$ EN &  Train (120k) &     124,770 & 379,546 & 319,484 \\
        KK $\leftrightarrow$ EN &  Train (220k) &     222,424 & 1,717,414 & 1,365,605 \\
        NE $\leftrightarrow$ EN &    Dev &       2,559 & 46,267 & 37,576 \\
        SI $\leftrightarrow$ EN &    Dev &       2,898 & 53,471 & 48,659 \\
        KK $\leftrightarrow$ EN &    Dev &       2,066 & 45,975 & 37,258 \\
        NE $\leftrightarrow$ EN &   Test &       2,835 & 51,455 & 43,802 \\
        SI $\leftrightarrow$ EN &   Test &       2,766 & 50,973 & 46,318 \\
        KK $\rightarrow$ EN &   Test &       1,000 & 20,376 & 15,943 \\
        EN $\rightarrow$ KK &   Test &        998 & 24,074 & 19,141 \\
\bottomrule
\end{tabular}
\caption{Number of sentences in raw corpora. The 120k and 220k training conditions for KK correspond to training KK$\leftrightarrow$EN models with/without an additional crawled corpus. The test sets for KK$\rightarrow$EN and EN$\rightarrow$KK are different from each other and mirror the released WMT19 data.}
\label{corpus-counts}
\end{table*}

\section{Related work}

Attempts to create unsupervised, morphologically-aware segmentations have often been derived from the Morfessor family of morphological segmentation tools \citep{virpioja2013morfessor}. In addition to extensions of Morfessor, such as \emph{Cognate Morfessor} \citep{gronroos-etal-2018-cognate}, \citet{ataman2017linguistically} and \citet{ataman-federico-2018-evaluation} introduced the LMVR model, derived from Morfessor \emph{FlatCat} \citep{gronroos-etal-2014-morfessor}, and applied it to NMT tasks on Arabic, Czech, German, Italian, Turkish and English, noting that LMVR outperforms a BPE baseline in CHRF3 and BLEU. Contrary to their results, however, \citet{toral-etal-2019-neural} find that using LMVR yielded mixed results: on a Kazakh-English translation task the authors observed marginal BLEU improvements over BPE, whereas for English-Kazakh, the authors reported LMVR to perform marginally worse than BPE in terms of CHRF3.

There have also been efforts to combine BPE with linguistically motivated approaches. For instance, \citet{huck-etal-2017-target} propose to combine BPE with various linguistic heuristics such as prefix, suffix, and compound splitting. 
The authors work with English-German and German-English tasks, and observe performance improvements of approximately 0.5 BLEU compared to a BPE-only baseline. 
As another example, \citet{weller-di-marco-fraser-2020-modeling} combine BPE with a full morphological analysis on the source and target sides of an English-German translation task, and report performance improvements exceeding 1 BLEU point over a BPE-only baseline.
 
Finally, even though \citet{sennrich-etal-2016-neural} originally only used the NMT training set to train their segmentation model, others have recently found benefit in adding monolingual data to the process. In particular, \citet{scherrer-etal-2020-university} used both SentencePiece and Morfessor as segmentation models on an Upper Sorbian--German translation task and found a monotonic increase in BLEU when the segmentation model was trained with additional data, while at the same time keeping the NMT training data constant.

\section{Experiments}
\label{ssec:accessibility}

\begin{table*}
\small
\centering
\begin{tabular}{ll}
\toprule
Segmentation & Sentence \\
\midrule
Original & The nation slowly started being centralized and during \\
SentencePiece & \_the \_n ation \_sl ow ly \_start ed \_being \_cent ral ized \_and \_d ur ing \\
% This abuse of math mode makes for better kerning around @
Subword-NMT & the n$@@$ ation s$@@$ low$@@$ ly star$@@$ ted being cen$@@$ tr$@@$ ali$@@$ z$@@$ ed and d$@@$ ur$@@$ ing \\ 
LMVR & the nation s +low +ly st +ar +ted be +ing c +ent +ral +ized and d +ur +ing \\ 
MORSEL & the nation s$@@$ low +ly start +ed being cen$@@$ tr$@@$ ali$@@$ z +ed and du$@@$ r +ing \\
\bottomrule
\end{tabular}
\caption{Examples of segmentation strategies and tokenization.}
\label{example-segmentations-table}
\end{table*}

To investigate the effect of subword segmentation algorithms on NMT performance, we train translation models using the Transformer architecture of \citet{vaswani2017attention}. 
We base our work on two recent datasets: FLoRes \citep{guzman-etal-2019-flores}, and select languages from the WMT 2019 Shared Task on News Translation \citep{barrault-etal-2019-findings}. 
Corpus statistics for all corpora can be found in Table~\ref{corpus-counts}.

The FLoRes dataset consists of two language pairs, English-Nepali and English-Sinhala. 
To add another lower-resourced language, we use the Kazakh-English translation data from  WMT19.
In terms of morphological typology, both Nepali and Sinhala are agglutinative languages \citep{prasain2011computational,priyanga-etal-2017-sinhala}, as is Kazakh \citep{kessikbayeva-cicekli-2014-rule}.

We conduct two sets of experiments on Kazakh to investigate how the amount of training data influences our results: first, we train only on the WikiTitles and News Commentary corpora (\texttt{train120k}), followed by another set of experiments (\texttt{train220k}) where we include the web crawl corpus prepared by Bagdat Myrzakhmetov of Nazarbayev University.
We also conducted experiments with Gujarati data from WMT19, but BLEU scores were too low to allow for meaningful analysis.
For our models, we generally follow the architecture and hyperparameter choices of the FLoRes Transformer baseline, except for setting \texttt{clip\_norm} to 0.1 and enabling FP16 training.

Despite the widespread use of auxiliary techniques such as back-translation we deliberately refrain from employing such techniques in this work. This is done to best isolate the effect of varying the subword segmentation algorithm, and to avoid the complexity of disentangling it from the effect of other factors. It should be noted, however, that such techniques were highly prevalent among of systems submitted to the KK$\leftrightarrow$EN WMT19 News Translation Shared Task: 64\% used back-translation, 61\% used ensembling, and 57\% employed extensive corpus filtering \citep{barrault-etal-2019-findings}. 

\subsection{Subword segmentation algorithms}

Below we describe our hyperparameter settings for the various subword segmentation algorithms.
Sinhala and Nepali are tokenized using the Indic NLP tokenizer \citep{kunchukuttan2020indicnlp}, whereas for English and Kazakh we use the Moses tokenizer \citep{koehn-etal-2007-moses}. Example segmentations from actual data can be seen in Table~\ref{example-segmentations-table}.

The segmentation methods we evaluate learn their subword vocabularies from frequency distributions of tokenized text. The exception to this is SentencePiece, whose subword units are learned from sentences, including whitespace. In the case of English and Kazakh, these sentences are untokenized whereas for Nepali and Sinhala, preprocessing with the Indic NLP tokenizer is applied following the approach of \citet{guzman-etal-2019-flores}. 

\subsubsection{Subword-NMT and SentencePiece}

As our baseline subword segmentation algorithm, we use the BPE implementation from Subword-NMT\footnote{\url{https://github.com/rsennrich/subword-nmt}}.
Throughout our experiments we use a joint vocabulary of the source and target and set the number of requested symbols to 5,000.
For SentencePiece, we use the default BPE implementation\footnote{\url{https://github.com/google/sentencepiece}} with a joint vocabulary size of 5,000 words. These choices are motivated by the general observation by \citet{sennrich-zhang-2019-revisiting} that lowering BPE size improves translation quality in ultra-low resource conditions, and the specific value of 5,000 was previously used by \citet{guzman-etal-2019-flores}. The same small vocabulary size has been used elsewhere in the low-resource NMT literature, for instance by \citet{roest2020morphological} while training NMT systems for Inuktitut. We also conducted a hyperparameter sweep for 2,500, 5,000, 7,500 and 10,000 merge operations, but noticed no improvement over the choice of 5,000 motivated by prior work.

\subsubsection{LMVR}

For LMVR \citep{ataman2017linguistically}, we utilize slightly modified versions of the sample scripts from the author's Github repository\footnote{\url{https://github.com/d-ataman/lmvr}}. Our main modification is tuning the \texttt{corpusweight} hyperparameter in the Morfessor Baseline \citep{virpioja2013morfessor} model used to seed the LMVR model. Tuning is performed by maximizing the F1 score for segmenting the English side of the training data, using the English word lists from the Morpho Challenge 2010 shared task \citep{kurimo2010proceedings} as gold standard segmentations. 
After tuning the Morfessor Baseline model, we train a separate LMVR model for each language in a language pair using a vocabulary size parameter of 2,500 per language. 

\subsubsection{MORSEL}

MORSEL \citep{lignos2010learning} provides linguistically-motivated unsupervised morphological analysis that has been shown to work effectively on small datasets \citep{chan2010investigating}.
While it provides derivations of morphologically complex forms via a combination of stems and affix rules, we modified it to provide a segmentation and then postprocessed its output to apply BPE to the stems to yield a limited-size vocabulary.

For example, on the English side of the NE-EN training data, MORSEL analyzes the word \emph{algebraic} as resulting from the stem \emph{algebra} being combined with the suffix rule \emph{+ic}.
A BPE model is trained on all of the stems in MORSEL's analysis, and when that is applied to the stem, it is segmented as \texttt{al@@ ge@@ br@@ a}.
The stem and suffix are combined using a special plus character to denote suffixation, so the final segmentation is \texttt{al@@ ge@@ br@@ a +ic}.
Tuning is performed as with LMVR, using the English word lists from the Morpho Challenge 2010 shared task \citep{kurimo2010proceedings} as a reference.
We adjust the number of BPE units learned from the stems to keep the total per-language vocabulary below 2,500.

\section{Results and analysis}

\begin{table}[tbh!]
\small
\centering
\begin{tabular}{llrr}
\toprule
Segm. method & BLEU & CHRF3\\
\midrule
EN-KK (train120k) & & \\
\midrule
LMVR &  1.00 $\pm$ 0.12 &  \textbf{21.98} $\pm$ 0.41 \\
MORSEL &  0.94 $\pm$ 0.11 &  \textbf{21.24} $\pm$ 0.89 \\
SentencePiece &  1.04 $\pm$ 0.09 &  \textbf{21.48} $\pm$ 0.47 \\
Subword-NMT &  \underline{\textbf{1.32}} $\pm$ 0.08 &  \underline{\textbf{22.12}} $\pm$ 0.28 \\
\midrule

EN-KK (train220k) & & \\
\midrule
LMVR & 1.82 $\pm$ 0.13 & \textbf{22.74} $\pm$ 0.84 \\
MORSEL &  \textbf{2.06} $\pm$ 0.11	& \textbf{22.88} $\pm$ 0.40 \\
SentencePiece &  \underline{\textbf{2.18}} $\pm$ 0.08 & \underline{\textbf{22.78}} $\pm$ 0.43 \\
Subword-NMT &  1.94 $\pm$ 0.22 & \textbf{22.62} $\pm$ 0.88 \\
\midrule
KK-EN (train120k) & & \\
\midrule
LMVR &  1.70 $\pm$ 0.07 & 23.72 $\pm$ 0.44 \\
MORSEL &  \textbf{2.62} $\pm$ 0.08	& \underline{\textbf{26.26}} $\pm$ 0.36 \\
SentencePiece &  2.34 $\pm$ 0.21 & 24.64 $\pm$ 0.81 \\
Subword-NMT &  \underline{\textbf{3.14}} $\pm$ 0.18 & \textbf{25.92} $\pm$ 0.54 \\

\midrule
KK-EN (train220k) & & \\
\midrule
LMVR &  9.42 $\pm$ 0.26 & 33.88 $\pm$ 0.76 \\
MORSEL &  \textbf{10.44} $\pm$ 0.48	& \textbf{34.58} $\pm$ 0.88 \\
SentencePiece &  \textbf{10.02} $\pm$ 0.29 & 33.50 $\pm$ 0.54 \\
Subword-NMT &  \underline{\textbf{10.68}} $\pm$ 0.34 & \underline{\textbf{35.52}} $\pm$ 0.41 \\
\midrule
EN-NE & & \\
\midrule
LMVR &  4.32 $\pm$ 0.04 & \textbf{31.00} $\pm$ 0.29 \\
MORSEL &  \textbf{4.38} $\pm$ 0.16 &	\textbf{31.28} $\pm$ 0.47 \\
SentencePiece &  \underline{\textbf{4.58}} $\pm$ 0.15 & \underline{\textbf{31.36}} $\pm$ 0.35 \\
Subword-NMT &  \textbf{4.42} $\pm$ 0.16 & 30.96 $\pm$ 0.34 \\
\midrule

NE-EN & & \\
\midrule
LMVR &  7.84 $\pm$ 0.11 & \textbf{34.10} $\pm$ 0.16 \\
MORSEL &  5.30 $\pm$ 0.30	& 28.18 $\pm$ 0.97 \\
SentencePiece &  \textbf{8.42} $\pm$ 0.23 & \underline{\textbf{34.40}} $\pm$ 0.73 \\
Subword-NMT &  \underline{\textbf{8.46}} $\pm$ 0.15 & \textbf{34.18} $\pm$ 0.13 \\
\midrule
EN-SI & & \\
\midrule
LMVR & \underline{\textbf{1.44}} $\pm$ 0.32 & \underline{\textbf{28.22}} $\pm$ 0.30 \\
MORSEL &  \textbf{1.12} $\pm$ 0.13	& 27.44 $\pm$ 0.34 \\
SentencePiece &  \textbf{1.08} $\pm$ 0.31 & \textbf{27.56} $\pm$ 0.43 \\
Subword-NMT &  0.88 $\pm$ 0.13 & 26.78 $\pm$ 0.51 \\
\midrule
SI-EN & & \\
\midrule
LMVR &  7.24 $\pm$ 0.22 & 32.16 $\pm$ 0.63 \\
MORSEL &  \underline{\textbf{7.78}} $\pm$ 0.16	& \textbf{34.32} $\pm$ 0.30 \\
SentencePiece &  \textbf{7.52} $\pm$ 0.08 & \textbf{33.58} $\pm$ 0.43 \\
Subword-NMT &  \textbf{7.76} $\pm$ 0.25 & \underline{\textbf{34.38}} $\pm$ 0.38 \\
\bottomrule
\end{tabular}
\caption{Mean and standard deviation of BLEU and CHRF3 across translation tasks and segmentation methods. Underlined values represent the highest mean scores. Bolded values are not significantly different ($p>0.05$) than the highest score as determined by Dunn's test.}
\label{raw-results-table}
\end{table}

\begin{table}[tbh]
\small
\centering
\begin{tabular}{lrr}
\toprule
Segmentation method &  BLEU &  CHRF3 \\
\midrule
        Subword-NMT &     6 &      6 \\
      SentencePiece &     6 &      6 \\
             MORSEL &     6 &      6 \\ 
              LMVR &     1 &      5 \\
\bottomrule
\end{tabular}
\caption{Number of times each segmentation method was or tied with the best-performing method under each metric, counted across all tasks.}
\label{tie-count-table}
\end{table}

\begin{figure*}[tb!]
\centering
\includegraphics[width=\linewidth]{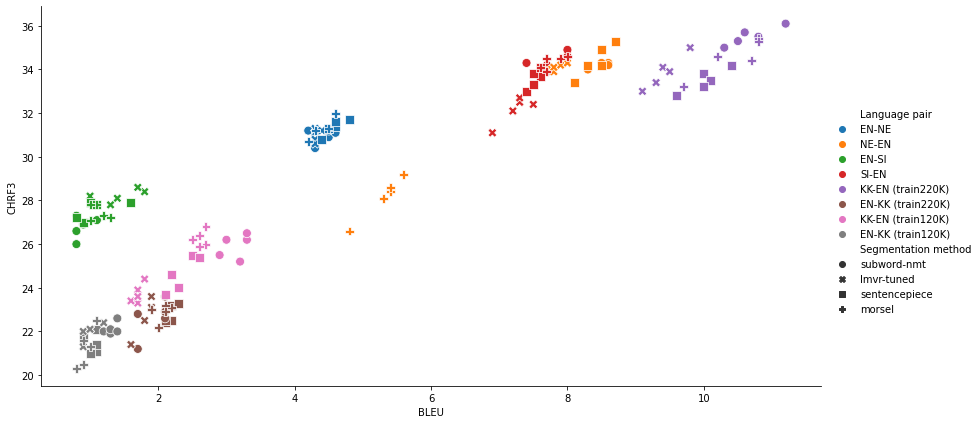}
\caption{CHRF3 vs. BLEU with different translation tasks indicated by color and segmentation by marker shape.}
\label{chrf3-vs-bleu}
\end{figure*}

Our experimental results can be seen in Table \ref{raw-results-table}.
All BLEU scores were computed using \texttt{sacrebleu}, and all CHRF3 scores using \texttt{nltk}.
Each row consists of the mean and standard deviation computed across 5 random seeds for each configuration.
We also plot the raw results in Figure \ref{chrf3-vs-bleu}.
Table~\ref{tie-count-table} gives counts for the number of times each segmentation approach was the top-performing one or statistically indistinguishable from it.
Table~\ref{dunns-table} in the appendix gives \emph{p}-values for all comparisons performed.

Overall, based on Tables \ref{raw-results-table} and \ref{tie-count-table}, no segmentation method seems to emerge as the clear winner across translation tasks, although BPE applied at the token (Subword-NMT) or sentence (SentencePiece) level performs well consistently.
Subword-NMT or SentencePiece perform best in 12 out of 16 cases (counting BLEU and CHRF3 for each translation task), while morphology-based methods rank best in 4 out of 16 cases.
In particular, we note that morphology-based methods seem to achieve or tie the best BLEU performance for translation tasks involving SI, and best CHRF3 performance for KK-EN with smaller training data (\texttt{train120k}) as well as EN-SI.
However, when using LMVR, we fail to find the significant gains in BLEU compared to BPE reported by \citet{ataman2017linguistically}.

Comparing our results to \citet{guzman-etal-2019-flores}, we note that the scores are similar, although not directly comparable as we report lowercased BLEU scores.\footnote{We lowercased all data in preprocessing because MORSEL and Morfessor, which LMVR is derived from, are designed to operate on lowercase inputs.}
They report EN-NE/NE-EN baseline BLEU scores of 4.3 and 7.6 using a single random seed, which are in line with our results in Table~\ref{raw-results-table}. For EN-SI/SI-EN, the authors report 1.2 and 7.2 BLEU, which likewise matches our findings. 
Even though our scores are low overall, they are as low as is to be expected using this approach, size of data, and languages. 
In order to compare our results to WMT19 participant systems, it is only meaningful to compare our system to baseline systems due to the widespread use of auxiliary training techniques, such as back-translation.
For instance, \citet{casas-etal-2019-talp} report baseline NMT scores of 2.32 on KK-EN and 1.42 on EN-KK, which are in line with our MORSEL and SentencePiece results on KK-EN, and Subword-NMT results on EN-KK in the \texttt{train120k} condition. 

\subsection{Modeling BLEU and CHRF3}\label{modeling}

\begin{table}[tbh]
\small
\centering
\resizebox{\linewidth}{!}{
\begin{tabular}{lrr}
\toprule
Pairwise comparison &      $\tau$ (BLEU) &    $\tau$ (CHRF3) \\
\midrule
SentencePiece - Subword-NMT &  -0.05 $\pm$ 0.08 &  -0.07 $\pm$ 0.20 \\
     MORSEL - Subword-NMT &  -0.12 $\pm$ 0.07 &  0.02 $\pm$ 0.18 \\
     LMVR - Subword-NMT &   -0.26 $\pm$ 0.06 &  -0.19 $\pm$ 0.21 \\
\bottomrule
\end{tabular}
}
\caption{Posterior means and standard deviations of $\tau_{m} - \tau_{Subword-NMT}$ (pairwise comparison with BPE) under the BLEU and CHRF3 models. Values are rounded to two decimal places.}
\label{pairwise-difference-stats-table}
\end{table}

Based on Figure \ref{chrf3-vs-bleu} and Tables~\ref{raw-results-table} and \ref{tie-count-table}, the BLEU and CHRF3 scores vary with both the translation task and segmentation method. Intuitively, the scores seem to cluster around a certain range for each translation task, and are perturbed slightly depending on the choice of segmentation method. To better disentangle the influence of these factors, we fit a Bayesian linear model to the experimental data, treating the final BLEU/CHRF3 score as a sum of a ``translation task effect'' $\eta$, a ``segmentation method effect'' $\tau$, and a translation task-specific noise term $\epsilon$.\footnote{In the appendix, Section~\ref{sec:bayesdetails} gives details of our model, and Table~\ref{full-posterior-inference-table} gives the point estimates of the posterior mean and standard deviation for $\eta$ and $\tau$.}
The $\eta$ and $\epsilon$ terms are estimated for each of the eight translation tasks (e.g. SI-EN and EN-SI are estimated separately), and $\tau$ is estimated for each of the four segmentation methods using results from all translation tasks.

To explicitly compare SentencePiece, LMVR and MORSEL to the Subword-NMT baseline, we also model the pairwise differences between each method's $\tau$-term and that of Subword-NMT. The posterior inferences for these quantities can be seen in Table~\ref{pairwise-difference-stats-table} and are plotted in the appendix.
For BLEU, the differences for LMVR are several standard deviations below 0, suggesting that it performs worse than the Subword-NMT baseline when accounting for all translation tasks. 
Similarly, MORSEL is almost 2 standard deviations away from 0, though its posterior interval does cover 0.
In both cases, the effect size is small, with a mean of -0.12 and -0.26 points of BLEU for MORSEL and LMVR, respectively.
The reliability of this difference also disappears for LMVR under the CHRF3 model, where no segmentation method's posterior mean is several standard deviations away from 0.

We hypothesize that this greater discrimination among methods when using BLEU may originate from the differences between how BLEU and CHRF3 operate.
Since CHRF3 is a character-level metric, it is less prone than BLEU to penalizing a given translation due to subword outputs that are \textit{almost} correct.
For instance, consider output of \texttt{do@@ gs} $\rightarrow$ \texttt{dogs} with \texttt{dog} as the reference; while CHRF3 awards credit for this as a partial match, BLEU treats it as entirely incorrect.
This further underscores our observation that segmentation methods perform inconsistently across experimental conditions.

\section{Conclusion and future work}

Contrary to our hypothesis about the usefulness of morphology-aware segmentation, we see no consistent advantage, and possibly a small disadvantage, to using LMVR or MORSEL in this resource-constrained setting.
By and large, our experiments and modeling show that no segmentation approach consistently achieves the best BLEU/CHRF3 across all translation tasks.
BPE remains a good default segmentation strategy, but it is possible that LMVR, MORSEL, or similar systems may show larger performance advantages for languages with specific morphological structures.

Consequently, we believe further work is needed to better understand when morphology-aware methods are most effective and to develop methods that provide a consistent advantage over BPE.
One such avenue of future work would be to broaden our analysis to more languages and include languages that are higher-resourced but morphologically rich and as well as ones that are lower-resourced but morphologically poor.
\citet{Ortega2021}, which we encountered during preparation of the final version of this paper, began to address these questions by comparing Morfessor with BPE and their own BPE variant on Finnish, Quechua and Spanish.

An alternative approach which we intend to pursue in future work is experimenting with supervised morphological segmenters or analyzers that can be efficiently developed even in lower-resourced settings.
Incorporating such ``gold standard'' segmentations may make it clearer whether the unsupervised morphological segmenters are capturing linguistically-relevant structure. 

Finally, there is the question of whether BPE can approximate a general representation for a language instead of converging on a corpus-specific set of subwords.
To test this, one can add monolingual data and train the BPE segmentation on that larger data set.
Ideally the new, ``enriched'' segmentations would depend less on the specific vocabulary of the training corpus.
As noted above, \citet{scherrer-etal-2020-university} observed this approach to be helpful in terms of BLEU. 
However, it remains unknown why the subwords derived from a larger corpus perform better, and whether better identification of morphological structure could be responsible.

We hope that this work and these ideas will catalyze further research, and that efficient methods for translating to and from lower-resourced languages can be developed as a result.

\bibliography{anthology,missing-from-anthology}
\bibliographystyle{acl_natbib}

\appendix

\section{Bayesian Linear Model Details}
\label{sec:bayesdetails}

Mathematically, our model can be expressed as: \begin{equation}
\phi_{lm} = \eta_l + \tau_m + \epsilon_l
\label{linear-model}
\end{equation} where $\phi_{lm} \in \{\text{BLEU, CHRF3}\}$, $\eta_l$ and $\tau_m$ represent the ``translation task effect" and ``segmentation method effect," and $\epsilon_l$ is a translation task-specific variance term.

To initialize our Bayesian linear model from Equation \ref{linear-model}, we set the following priors. For the BLEU model, $\eta_l \sim \mathcal{N}(4, 3)$ and $\tau_m \sim \mathcal{N}(0, 1)$. For the CHRF3 model, $\eta_l \sim \mathcal{N}(15, 7)$ and $\tau_m \sim \mathcal{N}(0, 1)$. The priors are the same regardless of translation task or segmentation method. For our noise terms, we use a $\epsilon_l \sim \text{HalfCauchy}(5)$ prior in all models.
Our rationale for these priors is that $\eta_l$ should place most of its probability mass within the observed range of BLEU/CHRF3, whereas $\tau_m$ should, \emph{a priori}, take on positive and negative values with equal probability, reflecting a lack of prior information.
All models are fit using PyMC3, and MCMC posterior inference performed using the No-U-Turn Sampler.

\begin{table}[tb]
\small
\centering
\resizebox{\linewidth}{!}{
\begin{tabular}{lrr}
\toprule
Segmentation method effect &      $\tau$ (BLEU) &    $\tau$ (CHRF3) \\
\midrule
               LMVR &  -0.09 $\pm$ 0.47 &  0.41 $\pm$ 0.50 \\
             MORSEL &  0.05 $\pm$ 0.47 &  0.63 $\pm$ 0.50 \\
      SentencePiece &   0.12 $\pm$ 0.47 &  0.53 $\pm$ 0.50 \\
        Subword-NMT &   0.17 $\pm$ 0.47 &  0.60 $\pm$ 0.50 \\
\midrule
Pairwise comparison &      $\tau$ (BLEU) &    $\tau$ (CHRF3) \\
\midrule
SentencePiece - Subword-NMT &  -0.05 $\pm$ 0.08 &  -0.07 $\pm$ 0.20 \\
     LMVR - Subword-NMT &  -0.26 $\pm$ 0.06 &  -0.19 $\pm$ 0.21 \\
     MORSEL - Subword-NMT &   -0.12 $\pm$ 0.07 &  0.02 $\pm$ 0.18 \\
\midrule
Translation task effect & $\eta$ (BLEU) & $\eta$ (CHRF3) \\
\midrule
        EN-KK (train120k) &  1.01 $\pm$ 0.47 &  21.16 $\pm$ 0.52 \\
        EN-KK (train220k) &  	1.94 $\pm$ 0.47 &  22.21 $\pm$ 0.51 \\
        EN-NE &  	4.36 $\pm$ 0.47 &  30.60 $\pm$ 0.50 \\
        EN-SI &  1.07 $\pm$ 0.47 &  26.95 $\pm$ 0.52 \\
        KK-EN (train120k) &  2.39 $\pm$ 0.48 &  24.58 $\pm$ 0.56 \\
        KK-EN (train220k) &  10.07 $\pm$ 0.48 &  33.81 $\pm$ 0.54 \\
        NE-EN &  7.41 $\pm$ 0.56 &  32.02 $\pm$ 0.82 \\
        SI-EN &  7.51 $\pm$ 0.47 &  33.05 $\pm$ 0.54 \\
\bottomrule
\end{tabular}}
\caption{Posterior means and standard deviations for $\tau$ and $\eta$ under the BLEU and CHRF3 models.}
\label{full-posterior-inference-table}
\end{table}

All posterior means for $\eta$ are close to the average BLEU/CHRF3 scores per translation task observed in Table~\ref{raw-results-table}, and fall between 1.01 and 10.07 for the BLEU model, and 21.16 and 33.81 for the CHRF3 model.
In contrast, the posterior means for $\tau$ are universally small: -0.09, 0.05,  0.12, and 0.17 for LMVR, MORSEL, SentencePiece and Subword-NMT, respectively, with a posterior standard deviation of 0.47. The $\tau$-terms under the CHRF3 model exhibit a similar pattern: 0.41, 0.63, 0.53, 0.60, with a posterior standard deviation of 0.50. Compared to the posterior standard deviation, as well as translation task effects $\eta$, the $\tau$-terms are practically 0. This, in conjunction with our analysis using Dunn's test, suggests that there is not a segmentation method that consistently works best across translation tasks.

\begin{figure*}[htb]
\centering
\includegraphics[width=\linewidth]{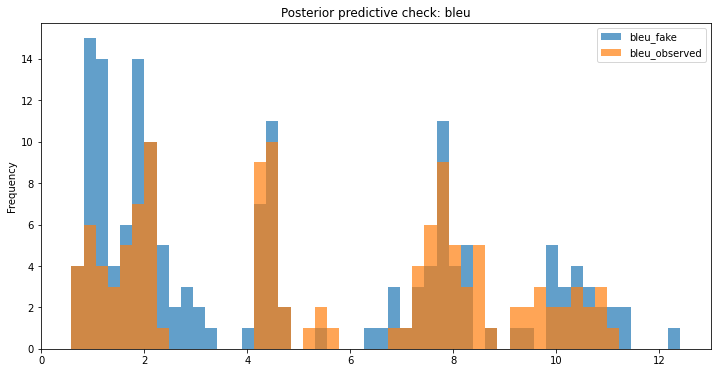}
\caption{Posterior predictive distribution of BLEU under the Bayesian linear model.}
\label{posterior-predictive-bleu}
\end{figure*}

\begin{figure*}[htb]
\centering
\includegraphics[width=\linewidth]{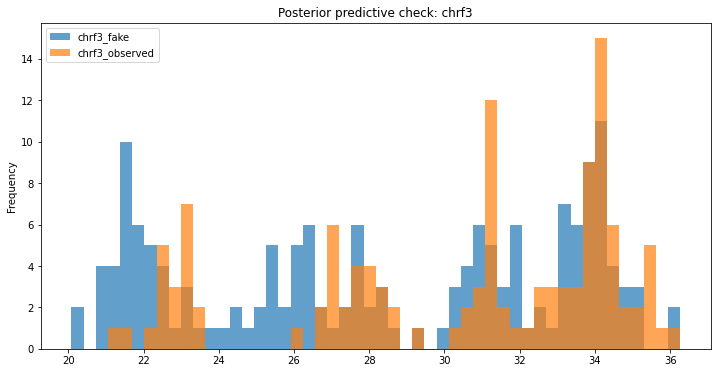}
\caption{Posterior predictive distribution of CHRF3 under the Bayesian linear model.}
\label{posterior-predictive-chrf3}
\end{figure*}

\begin{figure*}[tbh]
\centering
\includegraphics[width=\linewidth]{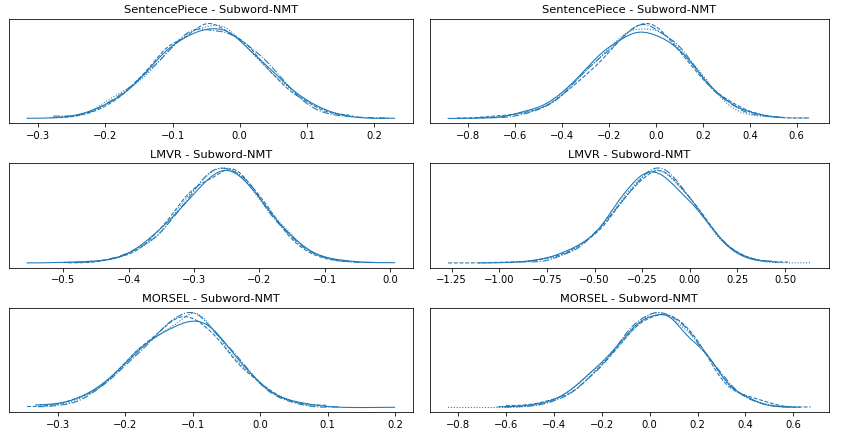}
\caption{Posterior distribution of pairwise differences $\tau_{m}-\tau_{\text{Subword-NMT}}$ in the BLEU model (left) and CHRF3 model (right). Note: $m \in \{\text{SentencePiece, LMVR, MORSEL}\}$}
\label{pairwise-differences-posterior}
\end{figure*}

Figures~\ref{posterior-predictive-bleu} and \ref{posterior-predictive-chrf3} show posterior predictive distributions for the BLEU and CHRF3 models.
Figure~\ref{pairwise-differences-posterior} shows the posterior distribution of pairwise differences between each of the other segmentation methods and Subword-NMT.

\begin{table*}[htb]
    \centering
    \begin{tabular}{llrr}
    \toprule
         Language pair & Segmentation method &  \emph{p}-value (BLEU) &  \emph{p}-value (CHRF3) \\
    \midrule
                 EN-NE &                LMVR &           \textbf{0.014} &            0.071 \\
                 EN-NE &              MORSEL &           0.057 &            0.466 \\
                 EN-NE &       SentencePiece &           1.000 &            1.000 \\
                 EN-NE &         Subword-NMT &           0.137 &            \textbf{0.046} \\
    \midrule
                 NE-EN &                LMVR &           \textbf{0.036} &            0.405 \\
                 NE-EN &              MORSEL &           \textbf{0.001} &            \textbf{0.002} \\
                 NE-EN &       SentencePiece &           0.872 &            1.000 \\
                 NE-EN &         Subword-NMT &           1.000 &            0.767 \\
    \midrule
                 EN-SI &                LMVR &           1.000 &            1.000 \\
                 EN-SI &              MORSEL &           0.246 &            \textbf{0.036} \\
                 EN-SI &       SentencePiece &           0.071 &            0.091 \\
                 EN-SI &         Subword-NMT &           \textbf{0.003} &   \textbf{0.000} \\
    \midrule
                 SI-EN &                LMVR &           \textbf{0.002} &   \textbf{0.001} \\
                 SI-EN &              MORSEL &           1.000 &            0.851 \\
                 SI-EN &       SentencePiece &           0.080 &            0.057 \\
                 SI-EN &         Subword-NMT &           0.850 &            1.000 \\
    \midrule
     KK-EN (train220k) &                LMVR &           \textbf{0.001} &   \textbf{0.009} \\
     KK-EN (train220k) &              MORSEL &           0.592 &            0.149 \\
     KK-EN (train220k) &       SentencePiece &           0.069 &            \textbf{0.002} \\
     KK-EN (train220k) &         Subword-NMT &           1.000 &            1.000 \\
    \midrule
     EN-KK (train220k) &                LMVR &           \textbf{0.002} &            0.788 \\
     EN-KK (train220k) &              MORSEL &           0.216 &            0.768 \\
     EN-KK (train220k) &       SentencePiece &           1.000 &            1.000 \\
     EN-KK (train220k) &         Subword-NMT &           \textbf{0.037} &            0.893 \\
    \midrule
     KK-EN (train120k) &                LMVR &           \textbf{0.000} &   \textbf{0.001} \\
     KK-EN (train120k) &              MORSEL &           0.140 &            1.000 \\
     KK-EN (train120k) &       SentencePiece &           \textbf{0.011} &   \textbf{0.026} \\
     KK-EN (train120k) &         Subword-NMT &           1.000 &            0.611 \\
    \midrule
     EN-KK (train120k) &                LMVR &           \textbf{0.008} &            0.872 \\
     EN-KK (train120k) &              MORSEL &           \textbf{0.001} &            0.068 \\
     EN-KK (train120k) &       SentencePiece &           \textbf{0.032} &            0.096 \\
     EN-KK (train120k) &         Subword-NMT &           1.000 &            1.000 \\
    \bottomrule
    \end{tabular}
    \caption{Dunn's test \emph{p}-values for BLEU and CHRF3. Boldface indicates statistical significance at the $\alpha=0.05$ level.}
    \label{dunns-table}
\end{table*}

\end{document}